\documentclass[letterpaper]{article}

\usepackage[preprint]{aaai2027}

\usepackage[hyphens]{url}
\usepackage{graphicx}
\usepackage{natbib}
\usepackage{caption}
\usepackage{amsmath}
\usepackage{amssymb}
\usepackage{mathtools}
\usepackage{algorithm}
\usepackage{algorithmic}
\usepackage{adjustbox}
\usepackage{placeins}

\usepackage{booktabs}
\usepackage{multirow}
\usepackage{xcolor}
\usepackage{fontawesome5}

\usepackage{xfp}

\newcommand{\R}{\mathbb{R}}
\newcommand{\E}{\mathbb{E}}
\newcommand{\N}{\mathcal{N}}
\newcommand{\D}{\mathcal{D}}
\newcommand{\G}{\mathcal{G}}
\newcommand{\A}{\mathcal{A}}
\newcommand{\Sspace}{\mathcal{S}}

\newcommand{\clip}{\operatorname{clip}}

\newcommand{\argmax}{\operatorname*{arg\,max}}

\newcommand{\epsg}{\varepsilon_\gamma}

\title{
PathBridger: Subgoal Bridges for Offline Goal-Conditioned Reinforcement Learning
}

\author{
    Soohyun Choi\textsuperscript{*},
    Seonvin Cho\textsuperscript{*},
    Songnam Hong\textsuperscript{\(\dagger\)}
}

\affiliations{
    Information and Intelligence Systems Laboratory (IISL)\\
    Department of Electronic Engineering, Hanyang University\\
    Seoul, Republic of Korea\\
    \texttt{\{petersun0221,seonbin0319,snhong\}@hanyang.ac.kr}\\[2pt]
    \textsuperscript{*}Equal contribution.
    \quad
    \textsuperscript{\(\dagger\)}Corresponding author.
}
\begin{document}

\maketitle

\definecolor{rangegray}{gray}{0.58}

\newcommand{\score}[3]{%
  \begingroup
  #3%
  \raisebox{0.10ex}{%
    {\normalfont\color{rangegray}%
     \fontsize{5.0pt}{5.0pt}\selectfont
     \kern0.12em
     $\pm$\kern0.08em
     \fpeval{round((#2-#1)/2,1)}}%
  }%
  \endgroup
}

\begin{abstract}
Offline goal-conditioned reinforcement learning (GCRL) aims to learn policies for reaching diverse goals entirely from fixed trajectory data. Long-horizon offline GCRL remains challenging because sparse goal-reaching signals must be propagated over many steps, while execution errors cannot be corrected through additional environment interaction. Existing methods address these challenges by improving long-range value estimation or reducing the effective decision horizon through subgoals, options, and action chunks. In several hierarchical methods, however, a selected subgoal specifies \emph{where to go}, while the intervening state-space path remains implicit in an endpoint-conditioned low-level policy. To address this interface, we propose \textbf{PathBridger}, a hierarchical offline GCRL method that explicitly connects subgoal selection to short-horizon execution. PathBridger constructs a state-space bridge toward the selected intermediate endpoint and decodes it into a short executable action chunk using an inverse dynamics model. Experiments across the evaluated OGBench tasks demonstrate strong aggregate performance, with particularly large gains on the multi-object Cube manipulation tasks. \textbf{\faGithub\ Code:}
\url{https://github.com/SChoish/PathBridger}

\end{abstract}

\section{Introduction}
\label{sec:introduction}

Reinforcement learning (RL) provides a general framework for learning sequential decision-making policies by maximizing cumulative rewards. In complex control problems, however, designing dense reward functions often requires substantial domain expertise and manual engineering, while misspecified proxy rewards can favor unintended shortcuts rather than the desired behavior~\cite{amodei2016concrete,hadfieldmenell2017inverse}. For many goal-reaching problems, it is considerably easier to specify a desired outcome, such as a target position, object configuration, or terminal observation, than to assign an informative reward to every intermediate state and action. Goal-conditioned reinforcement learning (GCRL) addresses this setting by learning a policy $\pi(a \mid s,g)$ that reaches goals $g$ drawn from a goal space, typically using sparse goal-reaching supervision~\cite{kaelbling1993learning,andrychowicz2017hindsight}.

Offline GCRL further aims to learn such goal-reaching policies entirely from fixed trajectory data, without additional environment interaction for training. This setting presents two closely related challenges. First, sparse goal-reaching signals must be propagated over long temporal horizons. For distant goals, the value differences induced by locally distinct decisions can become small relative to approximation and sampling errors, resulting in inaccurate value ordering and unreliable policy-improvement signals~\cite{park2023hiql,ahn2025ota}. Repeated temporal-difference backups can additionally accumulate estimation bias over the horizon~\cite{park2026transitive}. Second, long-range goal information must ultimately be translated into actions that are reliable under the offline data distribution. Errors caused by inaccurate values, policy extrapolation, or insufficient local coverage cannot be repaired through further data collection or online policy updates.

Prior work has addressed these challenges by improving long-range value estimation and introducing temporal abstraction. Option-based objectives and action-chunk methods reduce the number of value backups or policy decisions required over the same environment horizon~\cite{sutton1999options,li2025dqc,park2025mac}. Hierarchical GCRL decomposes long-horizon behavior into a high-level policy that selects intermediate subgoals and a low-level policy that acts toward them~\cite{park2023hiql,ahn2025ota}. In several hierarchical methods, the selected intermediate state is passed directly to an endpoint-conditioned low-level policy, leaving the intervening state sequence implicit at inference. Although such policies can accurately reach suitable nearby subgoals in some domains~\cite{ahn2025ota}, our controlled analyses suggest that endpoint-conditioned execution may not always be sufficient and that learned subgoal selection can remain a separate bottleneck. These observations motivate an explicit path-level interface between subgoal selection and action execution.

In this work, we introduce \textbf{PathBridger}, a hierarchical offline GCRL framework that explicitly connects subgoal selection to action execution. Given a current state \(s\) and goal \(g\), PathBridger selects an intermediate subgoal, constructs a state-space bridge toward it, and decodes the bridge into executable actions using an inverse-dynamics model.

Across the evaluated OGBench tasks~\cite{park2025ogbench}, PathBridger achieves the highest aggregate performance, with its clearest gains on multi-object Cube manipulation rather than uniform dominance across task families. Controlled bridge-removal and oracle-waypoint analyses further support the role of explicit bridge-based execution while suggesting that learned subgoal proposal and selection remain a limitation on Puzzle. A dataset-size ablation additionally finds that PBF attains the highest aggregate score at every tested dataset size.

Our main contributions are summarized as follows:
\begin{itemize}
    \item We identify an underexplored interface in hierarchical offline GCRL: in several endpoint-conditioned hierarchies, subgoal selection specifies the destination while leaving the intervening state-space path implicit.
    \item We propose PathBridger, which connects distributional endpoint proposal and transitive value-based selection to endpoint-pinned state-space bridge construction, inverse-dynamics action decoding, and receding-horizon execution. We instantiate PathBridger with Gaussian and rectified-flow endpoint proposers, yielding PathBridger-Gaussian (PBG) and PathBridger-Flow (PBF).
    \item We demonstrate the highest aggregate performance on the evaluated OGBench tasks, with particularly large gains on multi-object Cube manipulation, and use controlled bridge-removal and oracle-waypoint analyses to distinguish the role of bridge-based execution from limitations in learned subgoal selection.
\end{itemize}

\section{Related Work}
\label{sec:related-work}

\paragraph{Offline goal-conditioned reinforcement learning.}
Offline GCRL lies at the intersection of offline RL and goal-conditioned RL, aiming to learn reusable goal-reaching policies from fixed trajectory datasets~\citep{levine2020offline,kaelbling1993learning,andrychowicz2017hindsight}. Existing methods mainly differ in how they represent long-horizon reachability and extract policies from offline data. Goal-conditioned behavioral cloning (GCBC) directly imitates actions conditioned on hindsight-relabeled future goals and forms the basis of several goal-conditioned supervised learning approaches~\citep{ghosh2021learning,yang2022rethinking}. Goal-conditioned implicit V- and Q-learning (GCIVL and GCIQL) adapt in-sample offline value learning to state--goal pairs and extract policies using advantage-weighted or behavior-regularized objectives~\citep{kostrikov2022iql,park2023hiql,park2025ogbench}. Contrastive RL (CRL) learns behavioral goal-reaching values through contrastive classification, whereas quasimetric RL (QRL) and contrastive successor-feature methods learn directed temporal-distance structures for goal reaching~\citep{eysenbach2022contrastive,wang2023qrl,myers2024temporal}. OGBench provides a unified evaluation suite for these approaches across locomotion, manipulation, stitching, long-horizon reasoning, and high-dimensional observations~\citep{park2025ogbench}.

\paragraph{Hierarchical reinforcement learning and subgoal planning.}
Hierarchical RL represents temporally extended decisions through options, skills, macro-actions, or intermediate subgoals~\citep{sutton1999options}. In goal-conditioned settings, long-horizon tasks have also been addressed by planning over landmarks, replay-buffer graphs, or sequences of intermediate goals~\citep{savinov2018semiparametric,eysenbach2019sorb,nasiriany2019planning,li2022hierarchical}. HIQL develops a hierarchical approach for offline GCRL by extracting a high-level subgoal policy and a low-level action policy from a shared goal-conditioned value function~\citep{park2023hiql}. OTA studies the high-level policy bottleneck in this framework and introduces option-aware value backups to improve subgoal extraction over long horizons~\citep{ahn2025ota}.

\paragraph{Long-horizon value learning and action chunking.}
Temporal-difference learning propagates information through recursive value backups, which can accumulate estimation bias over long horizons. Multi-step returns and Monte Carlo objectives reduce the number of recursive backups, but introduce trade-offs involving off-policy bias, variance, and task-dependent backup horizons~\citep{park2025horizon}. TRL exploits the transitive structure of goal-conditioned values to perform divide-and-conquer updates over trajectory segments, reducing recursive value propagation without fixing a single multi-step horizon~\citep{park2026transitive}. A complementary approach reduces the effective horizon by treating temporally extended action sequences as the basic unit of policy or value learning. Q-chunking (QC) learns critics and policies over full action chunks, enabling multi-step value backups directly in the chunked action space~\citep{li2025actionchunking}. DQC decouples the critic and policy chunk lengths by distilling a partial-chunk critic from a longer action-chunk critic, preserving long-horizon value propagation while allowing shorter and more reactive execution~\citep{li2025dqc}. Related horizon-reduction methods combine multi-step targets with hierarchical policies and expressive behavioral models~\citep{park2025horizon}. In model-based offline RL, MAC uses action-chunk dynamics and action-chunk policies to reduce autoregressive model errors during long imaginary rollouts~\citep{park2025mac}. 

\paragraph{Trajectory generation and inverse dynamics.}
A complementary class of approaches models state or state--action sequences for planning and control. Trajectory Transformer models offline trajectories autoregressively, while Diffuser formulates planning as conditional denoising of trajectories~\citep{janner2021trajectory,janner2022diffuser}. Hierarchical and latent trajectory models, including TAP and hierarchical diffusion planning, reduce long-horizon planning complexity by generating compact latent plans or decomposing trajectories across multiple temporal scales~\citep{jiang2023tap,chen2024hierarchicaldiffusion}. Graph-based and model-based goal planners similarly search over intermediate states using learned reachability or dynamics models~\citep{eysenbach2019sorb,li2022hierarchical}. Inverse-dynamics models provide a mechanism for translating desired state transitions into actions and have been widely used in imitation learning and learning from observation~\citep{torabi2018bco,schmeckpeper2020learning}.

\section{Background}
\label{sec:background}

\subsection{Offline Goal-Conditioned Reinforcement Learning}
\label{sec:bg-gcrl}

We consider a Markov process with state space $\Sspace\subset\R^{d_s}$, action space $\A\subset\R^{d_a}$, dynamics $p(s'\mid s,a)$, and goal space $\G\subseteq\Sspace$. Thus, each goal $g\in\G$ is itself a target state. Offline GCRL assumes access to a fixed dataset $\D=\{\tau^{(i)}\}_{i=1}^{N}$ of trajectories $\tau=(s_0,a_0,s_1,a_1,\ldots,s_T)$ collected by one or more unknown behavior policies.

Let $\phi:\Sspace\rightarrow\R^{g_{\mathrm{rep}}}$ denote a task-dependent goal-representation map, and let $\rho$ be a metric on $\R^{g_{\mathrm{rep}}}$. We define the goal region, sparse binary reward, and first hitting time as
\begin{align*}
    B_{\epsilon}(g)
    &:=\left\{s\in\Sspace\,\middle|\,\rho\bigl(\phi(s),\phi(g)\bigr)\leq\epsilon\right\},\\
    r(s,g)
    &:=\mathbf{1}\left\{s\in B_{\epsilon}(g)\right\},\\
    \tau_g
    &:=\inf\left\{t\geq 0\,\middle|\,s_t\in B_{\epsilon}(g)\right\},
\end{align*}
where $\tau_g=\infty$ if the goal region is never reached. We adopt a hitting-time convention in which the process terminates upon its first entry into $B_{\epsilon}(g)$, so the goal reward is received at most once.

The objective and corresponding goal-conditioned value function are
\begin{align*}
    J(\pi)
    &=\E_{s_0,g,\pi,p}\!\left[\sum_{t=0}^{\infty}\gamma^t r(s_t,g)\right]\\
    &=\E_{s_0,g,\pi,p}\!\left[\gamma^{\tau_g}\mathbf{1}\{\tau_g<\infty\}\right],\\
    V^\pi(s,g)
    &=\E_{\pi,p}\!\left[\sum_{t=0}^{\infty}\gamma^t r(s_t,g)\,\middle|\,s_0=s\right]\\
    &=\E_{\pi,p}\!\left[\gamma^{\tau_g}\mathbf{1}\{\tau_g<\infty\}\,\middle|\,s_0=s\right].
\end{align*}
When rewards are sparse, hindsight relabeling uses future states from the same trajectory as goals, thereby constructing successful state--goal pairs directly from the offline dataset~\citep{andrychowicz2017hindsight}. Nevertheless, reaching distant goals still requires long-range value propagation, stitching across trajectory segments, or an explicit form of temporal abstraction.

\subsection{Transitive Structure in Goal Reaching}
\label{sec:bg-transitive}

Goal reaching admits a natural compositional structure through intermediate states. To state this structure precisely, we consider exact state goals, corresponding to the special case $B_0(g)=\{g\}$. In a deterministic environment, let $d^\star(s,g)$ denote the minimum number of transitions required to reach $g$ from $s$, with $d^\star(s,g)=\infty$ if $g$ is unreachable. This directed temporal distance satisfies $d^\star(s,g)\leq d^\star(s,z)+d^\star(z,g)$ for any intermediate state $z\in\Sspace$, with equality when $z$ lies on a shortest path~\citep{kaelbling1993learning,park2026transitive}.

Under the binary hitting reward, $V^\star(s,g)=\gamma^{d^\star(s,g)}$, with the convention $\gamma^\infty=0$. Since $0<\gamma<1$, the distance inequality yields the value-space relation
\begin{equation}
    V^\star(s,g)
    \geq
    V^\star(s,z)V^\star(z,g).
    \label{eq:transitive-value-inequality}
\end{equation}
Thus, $V^\star(s,z)V^\star(z,g)$ provides a compositional lower bound on $V^\star(s,g)$, with equality when $z$ lies on a shortest path. TRL turns this relation into a divide-and-conquer value-learning algorithm, thereby reducing the number of recursive value updates required for long-horizon goal-reaching tasks~\citep{park2026transitive}. In the tolerance-based formulation above, $B_\epsilon(g)$ remains the practical goal-success region, while the point-goal relation provides the structural motivation for transitive state-pair value composition.

\section{PathBridger}
\label{sec:pathbridger}

\begin{figure*}[t]
    \centering
    \includegraphics[width=.65\linewidth]{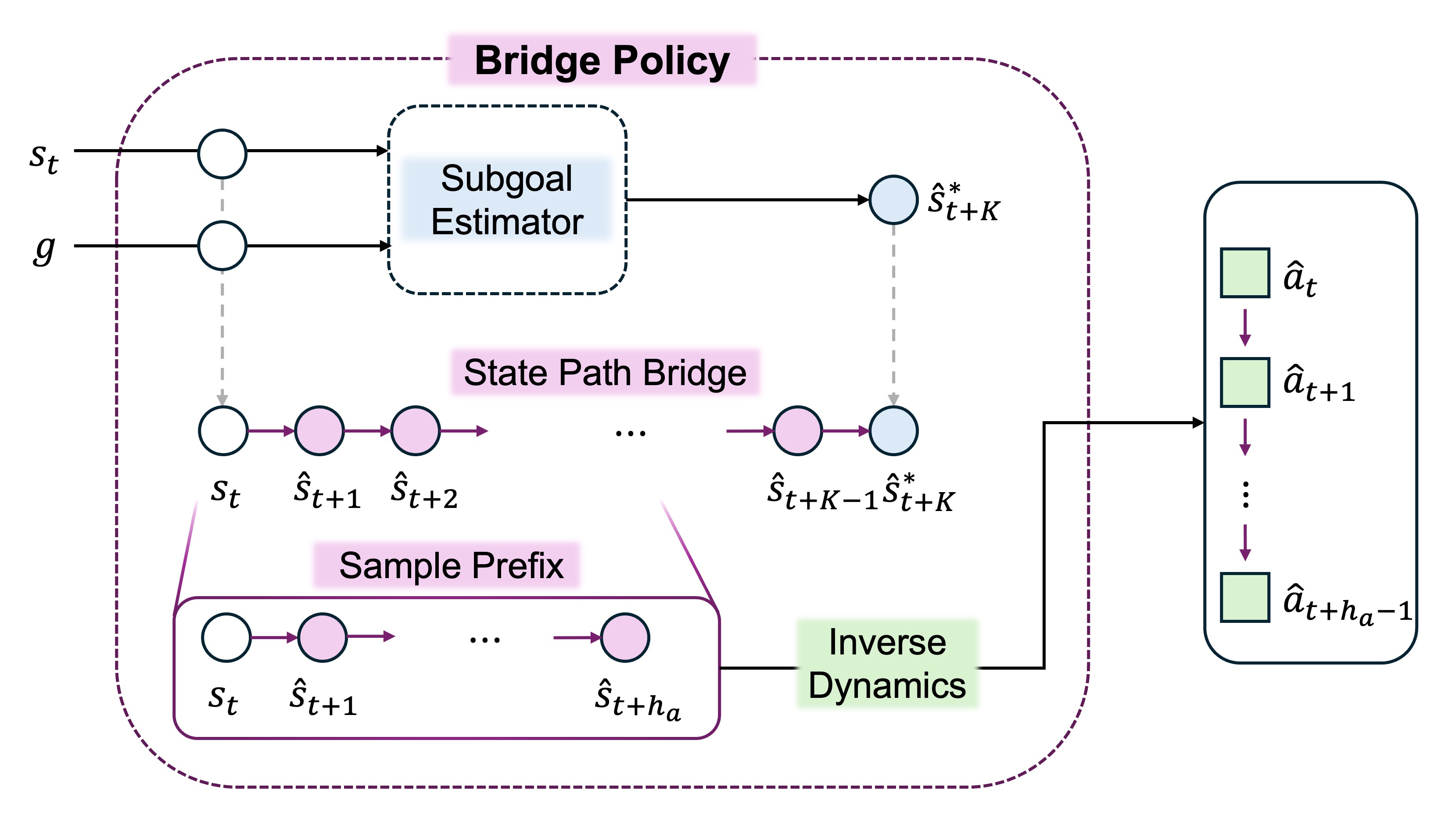
    }
    \caption{Overview of PathBridger. Candidate \(K\)-step endpoints are ranked using a transitive value score, and the selected endpoint \(z^\star\) is expanded into an endpoint-pinned state-space bridge. Inverse dynamics decodes an \(h_a\)-transition bridge prefix into an executable action chunk. PathBridger executes the chunk for up to \(h_a\) steps and then replans from the resulting state.}
    \label{fig:pathbridger-overview}
\end{figure*}

PathBridger is a \emph{bridge policy} that maps a current state and final goal to a short executable action chunk. Given \((s_t,g)\), it proposes candidate \(K\)-step endpoints, selects an endpoint using a transitive value score, expands the selected endpoint into an endpoint-pinned state-space bridge, and decodes the first \(h_a\) bridge transitions through inverse dynamics. The decoded action chunk is executed directly, after which PathBridger observes the resulting state and replans. In this design, transitive values determine \emph{where to bridge}, while the generated bridge specifies \emph{how to execute it}. Figure~\ref{fig:pathbridger-overview} provides an overview.

\subsection{Local Displacement Parameterization and Horizons}
\label{sec:disp}

Directly predicting future states in absolute coordinates can be poorly conditioned when the absolute-state scale is substantially larger than the local variation over a short trajectory window. PathBridger therefore represents endpoints and state paths in a displacement frame anchored at the current state. This removes the shared state-dependent offset, places prediction targets on the scale of local transitions, and improves numerical conditioning. We apply this parameterization in the Euclidean state representation used by the bridge model.

For a dataset window $(s_t^\D,\ldots,s_{t+K}^\D)$, define
\begin{equation}
\label{eq:disp-chart}
    \Delta_i^\D:=s_{t+i}^\D-s_t^\D,
    \qquad i=0,\ldots,K.
\end{equation}
Hence, $\Delta_0^\D=0$ and $\Delta_K^\D=s_{t+K}^\D-s_t^\D$. A predicted displacement path is mapped back to absolute states by
\begin{equation}
\label{eq:state-reconstruct}
    \hat s_{t+i}:=s_t+\hat\Delta_i,
    \qquad i=0,\ldots,K.
\end{equation}
The bridge horizon $K$ determines the endpoint prediction and state-space bridge, whereas the execution horizon $h_a\leq K$ determines the bridge prefix decoded by inverse dynamics and the number of actions executed before replanning.

\subsection{Transitive Value Learning}
\label{sec:transitive-value}

Motivated by the compositional relation in Equation~\eqref{eq:transitive-value-inequality}, PathBridger learns a bounded transitive value $V_\eta(s,g)\in(0,1)$ on the discounted hitting-time scale introduced in Section~\ref{sec:bg-transitive}. Let $\bar V_{\bar\eta}$ denote an exponentially averaged target network whose outputs are treated as fixed when constructing regression targets. Under deterministic dynamics, an observed $d$-step segment connecting $s$ to $g$ is a feasible path, so $d^\star(s,g)\leq d$ and therefore $\gamma^d\leq V^\star(s,g)$. We use these in-sample lower-bound targets to anchor short temporal distances and propagate them to longer distances through transitive composition.

Let $\ell_{\mathrm{BCE}}(v,y)$ denote binary cross-entropy between a prediction $v\in(0,1)$ and a soft target $y\in[0,1]$. Self-reachability and short in-trajectory pairs are anchored by
\begin{align*}
    \mathcal L_{\mathrm{self}}
    &=\E_{s\sim\D}\!\left[\ell_{\mathrm{BCE}}\!\left(V_\eta(s,s),1\right)\right],\\
    \mathcal L_{\mathrm{base}}
    &=\E_{\substack{\tau\sim\D,\ i<j\\j-i\leq H_b}}
      \!\left[\ell_{\mathrm{BCE}}\!\left(V_\eta(s_i^\D,s_j^\D),\gamma^{j-i}\right)\right].
\end{align*}

For a longer pair $(s_i^\D,g=s_j^\D)$, we sample an in-trajectory intermediate state $z=s_k^\D$, where $i<k<j$. For each constituent segment, we use its observed target inside the base horizon and the target-network prediction otherwise:
\begin{equation}
\label{eq:mixed-segment-value}
\widetilde V(s_a^\D,s_b^\D)
:=
\begin{cases}
    \gamma^{b-a}, & b-a\leq H_b,\\
    \bar V_{\bar\eta}(s_a^\D,s_b^\D), & b-a>H_b.
\end{cases}
\end{equation}
Using Equation~\eqref{eq:mixed-segment-value}, the corresponding transitive target is
\begin{equation}
\label{eq:value-tri-target-main}
    y_{\mathrm{tr}}(i,k,j)
    :=\widetilde V(s_i^\D,s_k^\D)\widetilde V(s_k^\D,s_j^\D).
\end{equation}
To approximate in-sample maximization over intermediate states, we use the expectile-weighted loss $\ell_{\mathrm{BCE}}^{\tau_V}(v,y):=|\tau_V-\mathbf 1\{v>y\}|\ell_{\mathrm{BCE}}(v,y)$, with $\tau_V\in[0.5,1)$. The target in Equation~\eqref{eq:value-tri-target-main} yields the transitive loss
\begin{equation}
\label{eq:value-tri-loss-main}
    \mathcal L_{\mathrm{tr}}
    =\E_{\substack{\tau\sim\D\\i<k<j}}
    \!\left[
        \ell_{\mathrm{BCE}}^{\tau_V}\!\left(
            V_\eta(s_i^\D,s_j^\D),
            y_{\mathrm{tr}}(i,k,j)
        \right)
    \right].
\end{equation}
Combining the transitive loss in Equation~\eqref{eq:value-tri-loss-main} with the self-reachability and short-range anchors, the complete objective is
\begin{equation}
\label{eq:value-total-loss-main}
    \mathcal L_V
    =
    \lambda_{\mathrm{self}}\mathcal L_{\mathrm{self}}
    +\lambda_{\mathrm{base}}\mathcal L_{\mathrm{base}}
    +\lambda_{\mathrm{tr}}\mathcal L_{\mathrm{tr}}.
\end{equation}
Following TRL, our implementation additionally applies distance-based re-weighting to emphasize shorter state--goal pairs, whose estimates serve as the constituents of longer transitive targets. Detailed implementation choices are provided in Appendix~\ref{app:implementation}.

The bridge policy ranks a candidate endpoint $z$ using the transitive score
\begin{equation}
\label{eq:subgoal-score}
    S_{\mathrm{tr}}(s,z,g)
    :=V_\eta(s,z)V_\eta(z,g).
\end{equation}
The first factor estimates reachability from the current state to the candidate endpoint, while the second estimates continuation from the candidate to the final goal.

\subsection{Bridge Policy}
\label{sec:bridge-policy}

The bridge policy maps $(s_t,g)$ to an executable action chunk through endpoint proposal, transitive selection, endpoint-pinned bridge construction, and inverse-dynamics decoding. The transitive value, endpoint proposer, bridge generator, and inverse-dynamics model are learned entirely from real offline trajectory segments; Algorithm~\ref{alg:pathbridger-training} summarizes the resulting training procedure.

\paragraph{Value-weighted endpoint proposal.}
For each dataset window, we pair the current state and $K$-step endpoint with a hindsight-relabeled future goal from the same trajectory. Let $p_\D^K(\Delta\mid s,g)$ denote the resulting empirical conditional distribution of $K$-step endpoint displacements. The bridge policy models $\Delta_K\sim q_\chi(\Delta_K\mid s,g)$ and maps the sampled displacement to the endpoint $z=s+\Delta_K$. We define the advantage-style centered continuation-value score
\begin{equation}
\label{eq:value-tilted-endpoint-distribution}
\begin{aligned}
    \delta_{\bar\eta}(s,g,\Delta)
    &:=\bar V_{\bar\eta}(s+\Delta,g)-\bar V_{\bar\eta}(s,g),\\
    q_{\mathrm{sg}}^\star(\Delta\mid s,g)
    &:=\frac{\exp\!\left(c_{\mathrm{sg}}\delta_{\bar\eta}(s,g,\Delta)\right)}
    {Z_{\bar\eta}(s,g)}
    p_\D^K(\Delta\mid s,g).
\end{aligned}
\end{equation}
Here, $c_{\mathrm{sg}}>0$ controls the strength of value tilting. Since $\bar V_{\bar\eta}(s,g)$ is independent of $\Delta$, the baseline term is absorbed into $Z_{\bar\eta}(s,g)$ and does not change the normalized conditional target. We retain this advantage-style centering in the unnormalized sample weights used for stochastic optimization to improve their numerical conditioning.

We consider two parameterizations of $q_\chi$. PathBridger-Gaussian (PBG) uses a diagonal Gaussian proposal, whereas PathBridger-Flow (PBF) represents the proposal as the terminal distribution of a conditional rectified flow and can capture multimodal endpoint distributions~\citep{lipman2023flowmatching}. Both variants are fit by reweighting dataset endpoint samples according to the unnormalized tilt in Equation~\eqref{eq:value-tilted-endpoint-distribution}; neither the empirical conditional density nor the normalizing constant $Z_{\bar\eta}(s,g)$ needs to be evaluated explicitly. Their concrete optimization objectives and goal-sampling rules are given in Appendix~\ref{app:implementation}.

Using the transitive score in Equation~\eqref{eq:subgoal-score}, the bridge policy samples and selects an endpoint at each replanning step by
\begin{equation}
\label{eq:best-of-n-subgoal}
\begin{aligned}
    \Delta_K^{(n)}&\sim q_\chi(\cdot\mid s_t,g),
    &&n=1,\ldots,N,\\
    z_n&=s_t+\Delta_K^{(n)},\\
    n^\star&=\argmax_{1\leq n\leq N}S_{\mathrm{tr}}(s_t,z_n,g),
    &&z^\star=z_{n^\star}.
\end{aligned}
\end{equation}

\paragraph{Endpoint-pinned bridge construction.}
For the selected endpoint, define $\Delta_K^\star:=z^\star-s_t$. A fixed monotone sequence $\{\alpha_i\}_{i=0}^K$, with $\alpha_0=0$ and $\alpha_K=1$, specifies a linear reference bridge. A learned residual deforms only the bridge interior:
\begin{equation}
\label{eq:residual-path}
    \hat\Delta_i
    =\alpha_i\Delta_K^\star
      +m_iR_\theta\!\left(s_t,\Delta_K^\star,\frac{i}{K}\right),
    \qquad i=0,\ldots,K.
\end{equation}
The fixed endpoint mask satisfies $m_0=m_K=0$. Consequently, $\hat\Delta_0=0$ and $\hat\Delta_K=\Delta_K^\star$, so the bridge is structurally pinned to both the current state and selected endpoint. The corresponding absolute states follow from Equation~\eqref{eq:state-reconstruct}.

For a real dataset window, the bridge is conditioned on the true $K$-step endpoint, while reconstruction is applied only to the first $h_a$ transitions used for action decoding:
\begin{equation}
\label{eq:bridge-loss-main}
\begin{aligned}
    \hat\Delta_i^\D
    &=\alpha_i\Delta_K^\D
      +m_iR_\theta\!\left(s_t^\D,\Delta_K^\D,\frac{i}{K}\right),\\
    \mathcal L_{\mathrm{bridge}}
    &=\E\!\left[
        \frac{1}{h_a}\sum_{i=1}^{h_a}
        \left\|\hat\Delta_i^\D-\Delta_i^\D\right\|_1
    \right].
\end{aligned}
\end{equation}
Since only this prefix is decoded into actions, restricting supervision to the execution-relevant transitions avoids unnecessary computation on the unexecuted remainder of the bridge. The interpolation sequence and endpoint mask used in our experiments are specified in Appendix~\ref{app:implementation}.

\paragraph{Inverse-dynamics decoding.}
The first $h_a+1$ bridge states define an $h_a$-transition prefix. Each consecutive state pair is decoded as
\begin{equation}
\label{eq:idm-decode}
    \hat a_{t+i}
    =I_\omega(\hat s_{t+i},\hat s_{t+i+1}),
    \qquad i=0,\ldots,h_a-1.
\end{equation}
The resulting bridge-induced action chunk is
\begin{equation}
\label{eq:bridge-action-chunk}
    \hat A_t(z^\star)
    :=(\hat a_t,\ldots,\hat a_{t+h_a-1})\in\A^{h_a}.
\end{equation}
The inverse-dynamics model is trained on real adjacent transitions:
\begin{equation}
\label{eq:idm-loss-main}
    \mathcal L_{\mathrm{idm}}
    =\E_{(s_t,a_t,s_{t+1})\sim\D}
    \!\left[\left\|I_\omega(s_t,s_{t+1})-a_t\right\|_2^2\right].
\end{equation}
PathBridger executes the decoded action chunk and replans from the resulting state, translating the selected endpoint into short-horizon actions through an explicit state-space path.

\begin{algorithm}[t]
\caption{PathBridger Training}
\label{alg:pathbridger-training}
\begin{algorithmic}[1]
\REQUIRE Offline dataset $\D$, bridge horizon $K$
\STATE Initialize $V_\eta$, target value $\bar V_{\bar\eta}$, endpoint proposer $q_\chi$, bridge generator $R_\theta$, and inverse dynamics $I_\omega$
\WHILE{not converged}
    \STATE Sample the required trajectory segments and hindsight goals from $\D$
    \STATE Update $V_\eta$ using \eqref{eq:value-total-loss-main} and update $\bar V_{\bar\eta}$ by exponential moving average
    \STATE Update $q_\chi$ to approximate the value-tilted endpoint distribution in \eqref{eq:value-tilted-endpoint-distribution}
    \STATE Update $R_\theta$ using \eqref{eq:bridge-loss-main}
    \STATE Update $I_\omega$ using \eqref{eq:idm-loss-main}
\ENDWHILE
\RETURN $V_\eta,q_\chi,R_\theta,I_\omega$
\end{algorithmic}
\end{algorithm}


\section{Experimental Results}
\label{sec:experimental-results}

\begin{table*}[t]
    \centering
    \caption{Average success rates on OGBench. Values are mean $\pm$ one standard deviation over four seeds. Each seed score averages evaluations at 800k, 900k, and 1M gradient steps. PBG and PBF use Gaussian and rectified-flow endpoint proposers, respectively. Bold indicates mean performance at or above 95\% of the best mean in each row.}
    \label{tab:overall_results_common_algorithms}
    \fontsize{7.6pt}{8.8pt}\selectfont
    \setlength{\tabcolsep}{3.4pt}
    \renewcommand{\arraystretch}{1.14}
    \begin{adjustbox}{max width=1.65\textwidth,center}
    \begin{tabular}{l*{9}{c}}
    \toprule
    \textbf{Environment}
    & \textbf{GCIVL}
    & \textbf{CRL}
    & \textbf{HIQL}
    & \textbf{OTA}
    & \textbf{TRL}
    & \textbf{HTVL}
    & \textbf{DQC}
    & \textbf{PBG}
    & \textbf{PBF} \\
    \midrule
    \texttt{antmaze-medium-navigate-v0}
    & \score{67.8}{78.6}{73.2}
    & \textbf{\score{93.2}{98.0}{95.6}}
    & \textbf{\score{95.5}{96.8}{96.2}}
    & \textbf{\score{95.8}{97.0}{96.4}}
    & \score{88.3}{91.9}{90.1}
    & \textbf{\score{92.2}{94.0}{93.1}}
    & \score{85.4}{88.5}{86.9}
    & \textbf{\score{94.7}{96.0}{95.3}}
    & \textbf{\score{95.8}{97.1}{96.4}} \\
    \texttt{antmaze-large-navigate-v0}
    & \score{18.0}{26.4}{22.2}
    & \textbf{\score{81.8}{91.8}{86.8}}
    & \textbf{\score{86.0}{93.6}{89.8}}
    & \textbf{\score{87.9}{93.9}{90.9}}
    & \score{39.5}{52.9}{46.2}
    & \score{85.6}{86.8}{86.2}
    & \score{67.5}{78.3}{72.9}
    & \score{82.4}{88.3}{85.3}
    & \score{78.2}{91.0}{84.6} \\
    \midrule
    \texttt{cube-single-play-v0}
    & \score{46.9}{50.5}{48.7}
    & \score{-0.9}{21.5}{10.3}
    & \score{12.3}{16.5}{14.4}
    & \score{16.7}{18.3}{17.5}
    & \score{89.7}{96.7}{93.2}
    & \score{83.4}{89.0}{86.2}
    & \score{72.5}{81.5}{77.0}
    & \score{91.3}{94.7}{93.0}
    & \textbf{\score{98.9}{99.9}{99.4}} \\
    \texttt{cube-double-play-v0}
    & \score{33.7}{42.5}{38.1}
    & \score{-0.8}{10.4}{4.8}
    & \score{6.6}{7.8}{7.2}
    & \score{0.8}{1.8}{1.3}
    & \score{13.2}{30.0}{21.6}
    & \score{27.5}{53.5}{40.5}
    & \score{15.1}{23.0}{19.0}
    & \score{73.8}{77.6}{75.7}
    & \textbf{\score{82.8}{87.3}{85.1}} \\
    \texttt{cube-triple-play-v0}
    & \score{0.4}{2.0}{1.2}
    & \score{-0.7}{3.5}{1.4}
    & \score{1.7}{4.0}{2.9}
    & \score{0.0}{1.0}{0.5}
    & \score{13.8}{14.8}{14.3}
    & \score{14.9}{30.1}{22.5}
    & \score{30.7}{32.9}{31.8}
    & \score{39.4}{50.9}{45.1}
    & \textbf{\score{75.3}{84.3}{79.8}} \\
    \midrule
    \texttt{puzzle-3x3-play-v0}
    & \score{5.0}{8.4}{6.7}
    & \score{2.4}{11.8}{7.1}
    & \score{8.8}{11.6}{10.2}
    & \score{51.6}{60.4}{56.0}
    & \textbf{\score{99.6}{100.0}{99.8}}
    & \score{19.6}{29.4}{24.5}
    & \textbf{\score{100.0}{100.0}{100.0}}
    & \score{25.2}{32.5}{28.9}
    & \score{66.0}{72.6}{69.3} \\
    \texttt{puzzle-4x4-play-v0}
    & \score{9.4}{11.6}{10.5}
    & \score{-0.1}{0.7}{0.3}
    & \score{3.5}{9.8}{6.7}
    & \score{8.7}{15.1}{11.9}
    & \score{32.0}{36.4}{34.2}
    & \score{1.5}{5.1}{3.3}
    & \textbf{\score{100.0}{100.0}{100.0}}
    & \score{1.1}{4.2}{2.7}
    & \score{72.0}{84.1}{78.1} \\
    \midrule
    \texttt{scene-play-v0}
    & \score{38.0}{48.2}{43.1}
    & \score{17.3}{19.1}{18.2}
    & \score{40.9}{45.8}{43.3}
    & \score{27.9}{34.4}{31.1}
    & \score{68.4}{78.6}{73.5}
    & \score{10.9}{14.9}{12.9}
    & \textbf{\score{86.3}{96.0}{91.1}}
    & \score{55.7}{63.3}{59.5}
    & \score{66.1}{72.1}{69.1} \\
    \midrule
    \textbf{Average}
    & $30.5$
    & $28.1$
    & $33.8$
    & $38.2$
    & $59.1$
    & $46.0$
    & $72.3$
    & $60.7$
    & \textbf{82.7} \\
    \bottomrule
    \end{tabular}
    \end{adjustbox}
\end{table*}

\begin{table}[t]
    \centering
    \caption{Estimated training time for 1M gradient steps on a single NVIDIA H200 GPU.}
    \label{tab:training-time}
    \fontsize{7.6pt}{8.8pt}\selectfont
    \setlength{\tabcolsep}{3.4pt}
    \renewcommand{\arraystretch}{1.14}
    \begin{tabular*}{\columnwidth}{@{\extracolsep{\fill}}lcccccccc}
    \toprule
    \textbf{Method}
    & \textbf{GCIVL}
    & \textbf{CRL}
    & \textbf{HIQL}
    & \textbf{OTA}
    & \textbf{TRL}
    & \textbf{DQC}
    & \textbf{PBG}
    & \textbf{PBF} \\
    \midrule
    \textbf{Time (h)} 
    & 0.94
    & 0.94
    & 1.40
    & 2.00
    & 1.34
    & 1.88
    & 1.20
    & 1.23 \\
    \bottomrule
    \end{tabular*}
\end{table}
\subsection{Experimental Setup}
\paragraph{Tasks.} We evaluate PathBridger on eight state-based OGBench tasks: two AntMaze navigation tasks, three Cube manipulation tasks, two Puzzle tasks, and one Scene task. We choose OGBench because it provides a standardized multi-goal evaluation across diverse offline GCRL domains, enabling consistent comparison with prior methods~\cite{park2025ogbench}. The selected tasks span long-horizon navigation, coordinated multi-object manipulation, and combinatorial control, allowing us to evaluate PathBridger across qualitatively different planning and execution challenges. We use the standard \texttt{navigate} datasets for AntMaze and the \texttt{play} datasets for the manipulation tasks.

\paragraph{Baselines and evaluation.} We compare PathBridger against GCIVL, CRL, HIQL, OTA, TRL, and DQC, and include Hierarchical Transitive Value Learning (HTVL) as a controlled baseline. For GCIVL, CRL, and HIQL, we use the official OGBench implementations without modification~\cite{park2025ogbench}. For DQC~\cite{li2025dqc}, we use the implementation of Song et~al.~\cite{song2026cgq}, which provides hyperparameters tuned for the 1M-transition setting considered here. For OTA and TRL, we follow the default settings from the original papers~\cite{ahn2025ota,park2026transitive} and tune only task-specific settings left unspecified by the authors. HTVL uses the same transitive value learner and Gaussian endpoint proposer as PBG but executes the selected endpoint with HIQL's endpoint-conditioned low-level actor. It therefore serves as a bridge-removal baseline for assessing bridge-based execution. For the main benchmark, all methods are trained for 1M gradient steps using four random seeds. Each seed score averages success over the five predefined OGBench goals at the 800k, 900k, and 1M checkpoints, and we report the mean and standard deviation across seeds. 

The experiments are run on NVIDIA H200 GPUs and Intel Xeon Platinum 8480C CPUs. Table~\ref{tab:training-time} reports the estimated steady-state training time for 1M gradient steps on a single H200 GPU. PBG and PBF require approximately \(1.20\) and \(1.23\) hours, respectively. Under the same setup, both variants train faster than HIQL, OTA, TRL, and DQC, while requiring more time than GCIVL and CRL.

\subsection{Main Benchmark Results}
Table~\ref{tab:overall_results_common_algorithms} reports the main OGBench results. PBF attains the highest overall average of 82.7, outperforming the strongest prior method by 10.4 points. Notably, PBG also achieves a strong average of 60.7, slightly outperforming TRL and substantially surpassing HIQL and OTA despite using a Gaussian endpoint proposer, indicating that the core PathBridger design remains competitive without the rectified-flow proposer. PathBridger's clearest gains occur on Cube, where PBF achieves the best result on all three tasks and the margin grows on the double- and triple-cube variants. On AntMaze, PathBridger matches the best result on \texttt{antmaze-medium} and remains competitive on \texttt{antmaze-large}. On Puzzle and Scene, PBF substantially outperforms HIQL and OTA but remains below DQC, with TRL also near-perfect on Puzzle-3x3.

\subsection{Ablation and Diagnostic Analyses}
\label{sec:ablation-analysis}
To better understand the source of PathBridger's benchmark gains and its remaining limitations on Puzzle, we conduct three complementary analyses: a bridge-removal ablation, an oracle-waypoint diagnostic, and a dataset-size sensitivity study.

\paragraph{Bridge-removal ablation.} HTVL shares PBG's transitive value learner, Gaussian endpoint proposer, and endpoint-selection procedure, but replaces bridge-based execution with HIQL's endpoint-conditioned low-level actor. HTVL improves the overall average from \(33.8\) for HIQL to \(46.0\), while PBG further reaches \(60.7\), a \(14.7\)-point gain over HTVL. Since HTVL and PBG differ primarily in endpoint execution, this gap supports the contribution of bridge construction and inverse-dynamics decoding.

\paragraph{Oracle-waypoint diagnostic on Puzzle.} We precompute a valid sequence of board configurations for each evaluation state--goal pair using lookup tables derived from the standard linear-algebraic formulation of Lights Out~\citep{anderson1998turning}. Each waypoint specifies the board configuration after the next prescribed button press in a valid solution. When a waypoint is supplied as the current task goal, the original long-horizon task is decomposed into shorter goal-reaching problems while each method retains its learned subgoal proposal and selection mechanism. When the same waypoint is supplied directly as the selected subgoal, learned proposal and selection are bypassed, leaving only execution. As shown in Table~\ref{tab:puzzle-oracle-waypoint}, PBG achieves \(98.7\) on both tasks when the waypoint is supplied as the selected subgoal, but achieves \(40.0\) and \(52.0\) when it is supplied as the task goal. This gap suggests that learned subgoal proposal and selection remain challenging even after the long-horizon task is externally decomposed. Under directly supplied subgoals, PBG also substantially outperforms HIQL, supporting the benefit of bridge-based execution over endpoint-conditioned low-level control. Together, these results identify learned subgoal proposal and selection as a key remaining bottleneck on Puzzle.

\begin{table}[t]
    \centering
    \caption{Mean success rates (\%) with oracle Puzzle waypoints. A waypoint is supplied either as the current task goal, retaining learned subgoal proposal and selection, or directly as the selected subgoal, evaluating execution alone. All results are averaged over two seeds.}
    \label{tab:puzzle-oracle-waypoint}
    \small
    \renewcommand{\arraystretch}{1.12}
    \begin{tabular*}{\columnwidth}{@{\extracolsep{\fill}}llcc@{}}
    \toprule
    \textbf{Environment}
    & \textbf{Waypoint role}
    & \textbf{PBG}
    & \textbf{HIQL} \\
    \midrule
    \multirow{2}{*}{\texttt{puzzle-3x3}}
    & Task goal
    & 40.0
    & 12.8 \\
    & Selected subgoal
    & \textbf{98.7}
    & 37.2 \\
    \addlinespace[2pt]
    \multirow{2}{*}{\texttt{puzzle-4x4}}
    & Task goal
    & 52.0
    & 7.2 \\
    & Selected subgoal
    & \textbf{98.7}
    & 2.4 \\
    \bottomrule
    \end{tabular*}
\end{table}

\begin{figure}[t]
    \centering
    \includegraphics[width=\columnwidth]{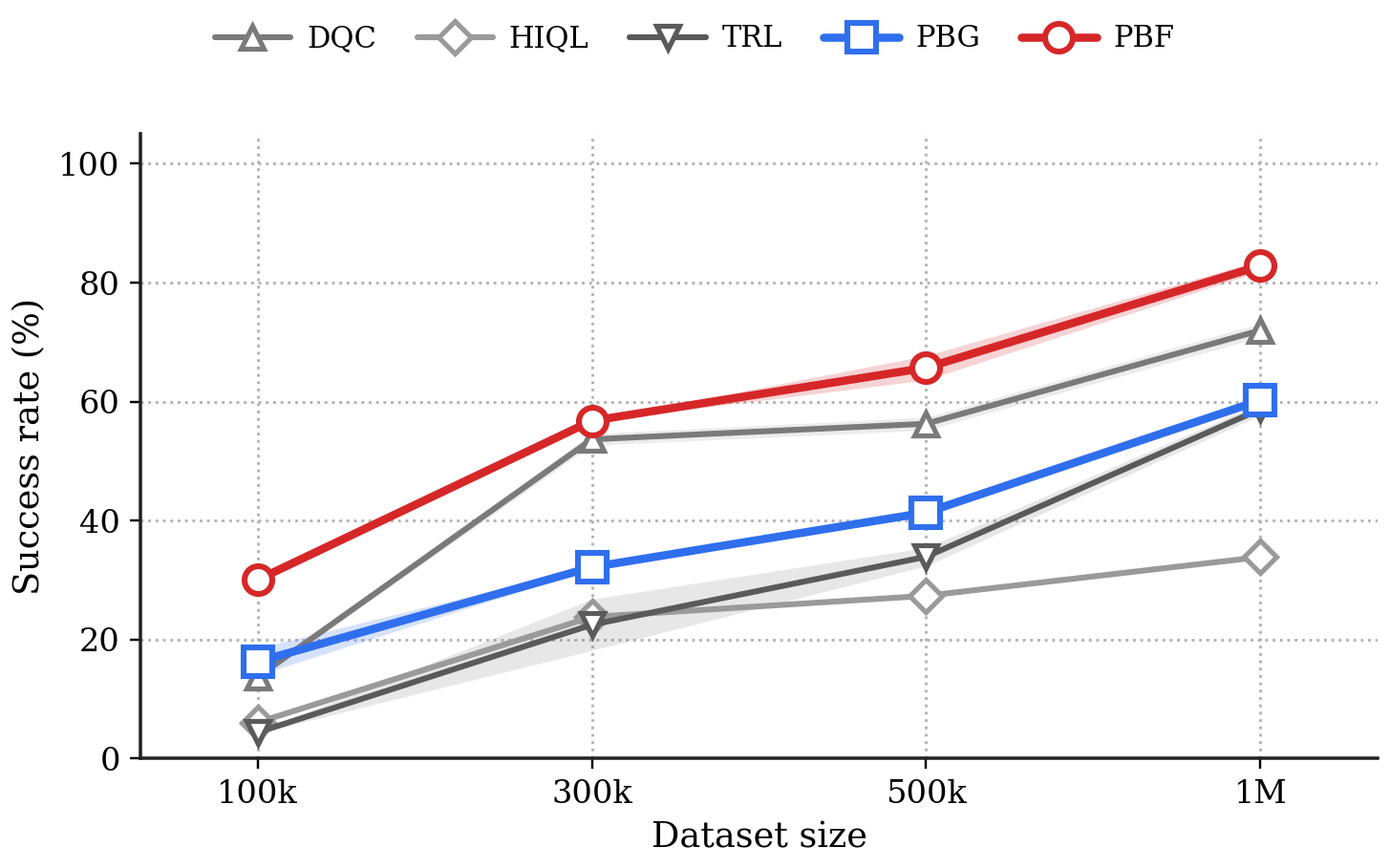}
    \caption{Dataset-size ablation averaged over all eight OGBench tasks. Each point reports the aggregate success rate with 100k, 300k, 500k, or 1M offline transitions while fixing training to 1M gradient steps.}
    \label{fig:dataset-size-scaling}
\end{figure}

\paragraph{Dataset-size sensitivity.} To examine sensitivity to the amount of offline data, we construct 100k-, 300k-, and 500k-transition subsets from each original 1M-transition dataset. We train PBF, PBG, DQC, HIQL, and TRL for 1M gradient steps at each data budget and evaluate them across all eight benchmark tasks. Figure~\ref{fig:dataset-size-scaling} reports the average success rate over these tasks. PBF attains the highest aggregate success rate at every tested dataset size, including 100k transitions. This result indicates that PBF's relative ranking is maintained across the evaluated data budgets.

\section{Conclusion}
\label{sec:conclusion}

We introduced PathBridger, a hierarchical offline GCRL framework that connects transitive value-based endpoint selection to endpoint-pinned state-space bridges and inverse-dynamics action decoding. Across eight state-based OGBench tasks, PBF achieves the highest aggregate performance, with its clearest gains on multi-object Cube manipulation.

Controlled analyses clarify both the benefit and the remaining limitation of this design. Removing the bridge lowers average performance, supporting the contribution of bridge-based execution beyond endpoint-conditioned low-level control. On Puzzle, supplying oracle waypoints directly as fixed subgoals yields nearly perfect success on both evaluated tasks, whereas performance remains substantially lower when learned subgoal proposal and selection are retained. These results suggest that PathBridger can reliably execute suitable intermediate states, while subgoal proposal and selection remain a key bottleneck under discrete, combinatorial board dynamics. Future work may therefore develop structure-aware or hybrid continuous--discrete subgoal models for such domains~\cite{gat2024discrete,holderrieth2025generator}.

\bibliography{pathbridger}

\clearpage
\appendix

\section{Implementation Details}
\label{app:implementation}

\subsection{Distance Reweighting for Transitive Values}
The value objective is computed from logits with binary cross-entropy. Following
TRL, we estimate a value-implied temporal distance and down-weight long
state--goal pairs:
\begin{align}
    \widehat d_\eta(s,g)
    &=
    \log_\gamma\!\left(
        \clip\!\left(\bar V_{\bar\eta}(s,g),\epsg,1\right)
    \right),\\
    w_\lambda(s,g)
    &=
    \left(1+\widehat d_\eta(s,g)\right)^{-\lambda}.
    \label{eq:distance_weight_supp}
\end{align}
We multiply the base and transitive losses by \(w_\lambda\), and stop gradients
through both the target values and the distance weights.

\subsection{Endpoint-Proposer Objectives}
For a dataset endpoint displacement \(\Delta_K^\D\), define
\begin{equation}
\begin{aligned}
    \delta_{\bar\eta}(s,g,\Delta)
    &=\bar V_{\bar\eta}(s+\Delta,g)-\bar V_{\bar\eta}(s,g),\\
    \widetilde w_\eta(s,g,\Delta_K^\D)
    &=\min\!\left\{
        w_{\max},
        \exp\!\left(c_{\rm sg}\delta_{\bar\eta}(s,g,\Delta_K^\D)\right)
    \right\}.
\end{aligned}
\label{eq:endpoint_weight_supp}
\end{equation}
We retain the centered form in the
unnormalized empirical weights to improve numerical conditioning. PBG minimizes
\begin{equation}
    \mathcal L_{\rm PBG}
    =
    -\E_{\D}\!\left[
        \widetilde w_\eta(s,g,\Delta_K^\D)
        \log q_\chi^{\rm G}(\Delta_K^\D\mid s,g)
    \right].
    \label{eq:pbg_loss_supp}
\end{equation}
For PBF, let \(x_0\sim\N(0,I)\), \(x_1=\Delta_K^\D\),
\(u\sim\operatorname{Unif}(0,1)\), and \(x_u=(1-u)x_0+ux_1\). Its weighted
flow-matching objective is
\begin{equation}
    \mathcal L_{\rm PBF}
    =
    \E\!\left[
        \widetilde w_\eta(s,g,x_1)
        \left\|
            v_\chi(s,g,x_u,u)-(x_1-x_0)
        \right\|_2^2
    \right].
    \label{eq:pbf_loss_supp}
\end{equation}
The target value and endpoint weight are treated as constants in both updates.

\subsection{Bridge and Sampling Details}
For \(i=0,\ldots,K\), the bridge interpolation and endpoint mask are
\begin{equation}
\begin{aligned}
    \alpha_i&=\left(\frac{i}{K}\right)^{0.8},
    &m_i&=\frac{i(K-i)}{K^2},\\
    \widehat\Delta_i
    &=\alpha_i\Delta_K
      +m_iR_\theta\!\left(s,\Delta_K,\frac{i}{K}\right).
\end{aligned}
\label{eq:bridge_schedule_supp}
\end{equation}
Because \(m_0=m_K=0\), the generated path is pinned to both endpoints.

The temperature \(T\) is used only for endpoint sampling at evaluation. PBG
samples
\begin{equation}
    \Delta^{(n)}
    =
    \mu_\chi(s,g)
    +T\,\sigma_\chi(s,g)\odot\epsilon^{(n)},
    \qquad
    \epsilon^{(n)}\sim\N(0,I).
    \label{eq:pbg_temperature}
\end{equation}
Thus, \(T=0\) evaluates the Gaussian mean. For PBF, the ODE is initialized
with \(x_0^{(n)}=T\epsilon^{(n)}\) and integrated using eight forward-Euler
steps. Training always uses \(x_0\sim\N(0,I)\), independently of the
evaluation temperature.

\section{Experimental Protocol and Hyperparameters}
\label{app:protocol_hyperparams}

\subsection{Evaluation Protocol}
Each OGBench environment provides five predefined evaluation goals. At each of
the 800k, 900k, and 1M checkpoints, we run 50 episodes per goal. We average the three checkpoints
within each seed and then report the mean and standard deviation across seeds.
Throughout the tables, \texttt{amm}, \texttt{aml}, \texttt{cs}, \texttt{cd},
\texttt{ct}, \texttt{p3}, \texttt{p4}, and \texttt{scene} denote
AntMaze-medium, AntMaze-large, Cube-single, Cube-double, Cube-triple,
Puzzle-3x3, Puzzle-4x4, and Scene, respectively.

\subsection{PathBridger Hyperparameters}

Tables~\ref{tab:appendix_shared_params} and~\ref{tab:pathbridger_env_complete} summarize the shared and environment-specific PathBridger hyperparameters, respectively. 
Unless otherwise specified, all loss coefficients are set to \(1\). In particular, the self-reachability, base, and transitive value losses are equally weighted, and the value, endpoint-proposal, bridge-reconstruction, and inverse-dynamics objectives are summed with unit coefficients. The environment-specific \(\lambda\) in Table~\ref{tab:pathbridger_env_complete} denotes the distance-reweighting exponent.

\begin{table}[t]
\centering
\small
\setlength{\tabcolsep}{4pt}
\renewcommand{\arraystretch}{1.10}
\begin{tabular}{@{}ll@{}}
\toprule
\textbf{Hyperparameter} & \textbf{Value} \\
\midrule
Learning rate & \(3\times10^{-4}\) \\
Optimizer & Adam \\
Batch size & 1024 \\
MLP hidden dimensions & \([512,512,512]\) \\
Activation & GELU \\
Target EMA rate & 0.005 \\
Layer normalization & Enabled \\
Base horizon \(H_b\) & 5 \\
Transitive expectile \(\tau_V\) & 0.7 \\
Distance-value clipping constant \(\varepsilon_\gamma\)
& \(10^{-6}\) \\
Execution horizon \(h_a\) & 5 \\
Endpoint-weight cap \(w_{\max}\) & 5.0 \\
Critic goal-sampling probabilities & \((0,1,0,0)\) \\
Endpoint goal-sampling probabilities & \((0,0,1,0)\) \\
Flow steps (PBF) & 8 \\
\bottomrule
\end{tabular}
\caption{Shared PathBridger hyperparameters. Goal-sampling probabilities are ordered as \((p_{\rm cur},p_{\rm geom},p_{\rm traj},p_{\rm rand})\).}
\label{tab:appendix_shared_params}
\end{table}

\begin{table*}[t]
\centering
\small
\setlength{\tabcolsep}{2.1pt}
\renewcommand{\arraystretch}{1.08}
\begin{tabular}{@{}l*{12}{c}@{}}
\toprule
& \multicolumn{6}{c}{\textbf{PBF}}
& \multicolumn{6}{c}{\textbf{PBG}} \\
\cmidrule(lr){2-7}
\cmidrule(lr){8-13}
\textbf{Environment}
& \(K\) & \(\gamma\) & \(c_{\rm sg}\) & \(\lambda\) & \(N\) & \(T\)
& \(K\) & \(\gamma\) & \(c_{\rm sg}\) & \(\lambda\) & \(N\) & \(T\) \\
\midrule
\texttt{amm}   & 25 & 0.99  & 10 & 0   & 2  & 0.25 & 25 & 0.99  & 10 & 0   & 1  & 0 \\
\texttt{aml}   & 25 & 0.995 & 10 & 0   & 16 & 0.5  & 25 & 0.995 & 10 & 0   & 1  & 0 \\
\texttt{cs}    & 40 & 0.99  & 5  & 0.7 & 1  & 0    & 25 & 0.99  & 10 & 0.7 & 1  & 0 \\
\texttt{cd}    & 40 & 0.99  & 10 & 1   & 2  & 0.25    & 25 & 0.99  & 10 & 1   & 1  & 0 \\
\texttt{ct}    & 40 & 0.995 & 10 & 1   & 1  & 0 & 25 & 0.995 & 10 & 1   & 1  & 0 \\
\texttt{p3}    & 25 & 0.99  & 10 & 0.5 & 32 & 1    & 40 & 0.99  & 10 & 0.5 & 2  & 0.25 \\
\texttt{p4}    & 25 & 0.99  & 10 & 2   & 32 & 1    & 40 & 0.995 & 10 & 2   & 16 & 0.5 \\
\texttt{scene} & 25 & 0.99  & 5  & 1   & 16 & 0.5  & 40 & 0.99  & 5  & 1   & 16 & 0.5 \\
\bottomrule
\end{tabular}
\caption{Environment-specific PathBridger hyperparameters.}
\label{tab:pathbridger_env_complete}
\end{table*}

\section{Baseline Details}
\label{app:baselines}

\subsection{Objectives}
GCIVL, CRL, and HIQL use the official OGBench implementations without
modification~\citep{park2025ogbench}. OTA and TRL follow their released
implementations and default objectives~\citep{ahn2025ota,park2026transitive}.
For DQC, we use the implementation of Song et~al.~\citep{song2026cgq} with
the DQC objectives~\citep{li2025dqc}. We summarize the objectives used in
our experiments and refer to the original papers for full derivations.

Let
\begin{align}
    \ell_{\rm exp}^{\tau}(u)
    &=\left|\tau-\mathbf 1\{u<0\}\right|u^2,\\
    \ell_{\rm quant}^{\tau}(u)
    &=\left|\tau-\mathbf 1\{u<0\}\right||u|,\\
    \ell_{\rm BCE}(p,y)
    &=-y\log p-(1-y)\log(1-p),\\
    \bar w_\beta(A)
    &=\min\{\exp(\beta A),w_{\max}\}.
\end{align}
We write \(r_-(s,g)=-\mathbf 1\{s\notin B_\epsilon(g)\}\), and bars denote
target networks.

\paragraph{GCIVL~\citep{kostrikov2022iql,park2025ogbench}.}
GCIVL learns an action-free state--goal value and extracts a flat policy by
advantage-weighted regression:
\begin{align}
    \mathcal L_V^{\rm GCIVL}
    &=
    \E\!\left[
        \ell_{\rm exp}^{\kappa}\!\left(
            r_-(s,g)+\gamma\bar V(s',g)-V(s,g)
        \right)
    \right],\\
    \mathcal L_\pi^{\rm GCIVL}
    &=
    -\E\!\left[
        \bar w_\beta\!\left(V(s',g)-V(s,g)\right)
        \log\pi(a\mid s,g)
    \right].
    \label{eq:gcivl_supp}
\end{align}

\paragraph{CRL~\citep{eysenbach2022contrastive,park2025ogbench}.}
Let \(f_\theta(s,a,g)\) denote the contrastive logit, let \(g^+\) be sampled from the future of \((s,a)\), let \(g^-\) be sampled independently from the dataset, and let \(\sigma\) denote the logistic sigmoid. CRL minimizes
\begin{equation}
\begin{aligned}
    \mathcal L_f^{\rm CRL}
    ={}&-\E\!\left[\log\sigma(f_\theta(s,a,g^+))\right]\\
       &-\E\!\left[\log\!\left(1-\sigma(f_\theta(s,a,g^-))\right)\right].
\end{aligned}
\label{eq:crl_critic_supp}
\end{equation}
The policy objective is
\begin{equation}
    \mathcal L_\pi^{\rm CRL}
    =-\E\!\left[
        f_\theta(s,\mu_\psi(s,g),g)
        +\alpha\log\pi_\psi(a\mid s,g)
    \right].
    \label{eq:crl_actor_supp}
\end{equation}
The implementation uses in-batch negatives and twin contrastive critics.

\paragraph{HIQL~\citep{park2023hiql}.}
HIQL uses the GCIVL value loss and extracts a high-level subgoal policy and a low-level action policy. For
\(z=s_{t+K}\),
\begin{align}
    A^h(s_t,z,g)&=V(z,g)-V(s_t,g),\\
    A^\ell(s_t,s_{t+1},z)&=V(s_{t+1},z)-V(s_t,z),
\end{align}
and
\begin{align}
    \mathcal L_{\pi^h}^{\rm HIQL}
    &=-\E\!\left[
        \bar w_{\beta_H}(A^h)\log\pi^h(z\mid s_t,g)
    \right] \nonumber,\\
    \mathcal L_{\pi^\ell}^{\rm HIQL}
    &=-\E\!\left[
        \bar w_{\beta_L}(A^\ell)\log\pi^\ell(a_t\mid s_t,z)
    \right].
    \label{eq:hiql_supp}
\end{align}
We suppress the learned subgoal representation in the notation.

\paragraph{OTA~\citep{ahn2025ota}.}
OTA retains HIQL's two-level policy extraction but trains a separate
option-aware high-level value. Let \(s^\Omega\) be the state reached after an
\(n\)-step dataset option or upon reaching \(g\). The low-level value uses the
GCIVL loss, while the high-level value minimizes
\begin{equation}
    \mathcal L_{V^h}^{\rm OTA}
    =
    \E\!\left[
        \ell_{\rm exp}^{\kappa}\!\left(
            r_-(s^\Omega,g)
            +\gamma\bar V^h(s^\Omega,g)
            -V^h(s,g)
        \right)
    \right].
    \label{eq:ota_value_supp}
\end{equation}
The actor losses follow Equation~\eqref{eq:hiql_supp}, using the separate
high-level value \(V^h\) and low-level value \(V^\ell\).

\paragraph{TRL~\citep{park2026transitive}.}
For an in-trajectory triplet \(i<k<j\), define
\begin{equation}
    \widetilde Q_{ab}
    =
    \begin{cases}
        \gamma^{b-a}, & b-a\leq 1,\\
        \bar Q(s_a,a_a,s_b), & b-a>1,
    \end{cases}
    \qquad
    y_{ikj}=\widetilde Q_{ik}\widetilde Q_{kj}.
\end{equation}
With
\begin{equation}
    w_\lambda(s_i,s_j)
    =
    \left(
        1+\log_\gamma
        \clip(Q(s_i,a_i,s_j),\epsg,1)
    \right)^{-\lambda},
\end{equation}
TRL minimizes
\begin{equation}
\begin{aligned}
    \mathcal L_Q^{\rm TRL}
    =\E\!\Big[
        &w_\lambda(s_i,s_j)
        \left|\kappa-\mathbf 1\{Q(s_i,a_i,s_j)>y_{ikj}\}\right|\\
        &\cdot\ell_{\rm BCE}\!\left(Q(s_i,a_i,s_j),y_{ikj}\right)
    \Big].
\end{aligned}
\label{eq:trl_supp}
\end{equation}
The default policy uses DDPG+BC,
\begin{equation}
    \mathcal L_\pi^{\rm TRL}
    =
    -\E\!\left[
        Q(s,a^\pi,g)+\alpha\log\pi(a\mid s,g)
    \right],
\end{equation}
while the Puzzle tasks use rejection sampling from a behavioral policy, following the original implementation.

\paragraph{DQC~\citep{li2025dqc}.}
DQC uses a full critic chunk
\(A_t^h=(a_t,\ldots,a_{t+h-1})\) and a shorter execution chunk
\(A_t^{h_a}=(a_t,\ldots,a_{t+h_a-1})\), where \(h_a<h\). Let
\(R_t^h(g)=\sum_{\ell=0}^{h-1}\gamma^\ell r(s_{t+\ell},g)\). Its three value
losses are
\begin{align}
    \mathcal L_Q^{\rm DQC}
    &=\E\!\left[
        \left(
            Q_\phi(s_t,A_t^h,g)
            -R_t^h(g)
            -\gamma^h\bar V_\xi(s_{t+h},g)
        \right)^2
    \right],\\
    \mathcal L_{Q^P}^{\rm DQC}
    &=\E\!\left[
        \ell_{\rm exp}^{\kappa_d}\!\left(
            \bar Q_\phi(s_t,A_t^h,g)
            -Q_\psi^P(s_t,A_t^{h_a},g)
        \right)
    \right],\\
    \mathcal L_V^{\rm DQC}
    &=\E\!\left[
        \ell_{\rm quant}^{\kappa_b}\!\left(
            \bar Q_\psi^P(s_t,A_t^{h_a},g)
            -V_\xi(s_t,g)
        \right)
    \right].
    \label{eq:dqc_values_supp}
\end{align}
The partial critic optimistically distills the value of full chunks sharing the
same prefix. DQC also trains a state-conditioned flow behavior model over
partial chunks. With \(x_0\sim\N(0,I)\), \(x_1=A_t^{h_a}\),
\(u\sim\operatorname{Unif}(0,1)\), and \(x_u=(1-u)x_0+ux_1\),
\begin{equation}
    \mathcal L_{\rm flow}^{\rm DQC}
    =
    \E\!\left[
        \left\|
            v_\beta(s_t,x_u,u)-(x_1-x_0)
        \right\|_2^2
    \right].
    \label{eq:dqc_flow_supp}
\end{equation}
At inference, DQC samples \(N\) partial chunks from this behavior model and
selects
\begin{equation}
    A_t^\star
    =
    \argmax_{1\leq n\leq N}
    Q_\psi^P(s_t,A_t^{(n)},g),
    \qquad
    A_t^{(n)}\sim\pi_\beta(\cdot\mid s_t).
    \label{eq:dqc_policy_supp}
\end{equation}
Thus, the behavior model is not goal-conditioned; goal-directed improvement is
performed by the partial critic.

\paragraph{HTVL.}
HTVL uses the same transitive value objective, Gaussian endpoint proposer, and endpoint-selection rule as PBG, but replaces bridge construction and inverse-dynamics decoding with HIQL's endpoint-conditioned low-level actor. For a future-state
subgoal \(z\),
\begin{align}
    A_{\rm HTVL}^\ell
    &=V_\eta(s_{t+1},z)-V_\eta(s_t,z),\\
    \mathcal L_{\pi^\ell}^{\rm HTVL}
    &=-\E\!\left[
        \bar w_{\beta_L}(A_{\rm HTVL}^\ell)
        \log\pi^\ell(a_t\mid s_t,z)
    \right].
    \label{eq:htvl_supp}
\end{align}
At inference, HTVL samples and ranks endpoints exactly as PBG and directly
passes the selected endpoint to the low-level actor.

\subsection{Baseline Hyperparameters}
\label{sec:baseline-hyperparameters}

Table~\ref{tab:appendix_baseline_shared} summarizes the optimization,
architecture, and goal-sampling settings shared across the baseline
algorithms. Unless otherwise specified, all baselines use the default
configuration in this table. Method-specific settings for OTA, TRL, and
DQC are reported subsequently.

\paragraph{Shared settings.}
The goal-sampling probability vectors are ordered as
\((p_{\rm cur},p_{\rm geom},p_{\rm traj},p_{\rm rand})\).

\begin{table}[ht]
    \centering
    \small
    \setlength{\tabcolsep}{5pt}
    \renewcommand{\arraystretch}{1.14}
    \begin{tabular}{@{}ll@{}}
    \toprule
    \textbf{Hyperparameter} & \textbf{Value} \\
    \midrule

    Learning rate
        & \(3\times10^{-4}\) \\
    \addlinespace[0.5ex]

    Optimizer
        & Adam \\
    \addlinespace[0.5ex]

    Batch size
        & \raisebox{-0.45\height}{%
            \shortstack[l]{%
                1024 (default)\\
                256 for DQC and OTA\(\dagger\)
            }} \\
    \addlinespace[0.7ex]

    MLP hidden dimensions
        & \raisebox{-0.45\height}{%
            \shortstack[l]{%
                \([512,512,512]\) (default)\\
                \([512,512,512,512]\) for DQC
            }} \\
    \addlinespace[0.7ex]

    Target EMA rate
        & 0.005 \\
    \addlinespace[0.5ex]

    Policy goal-sampling probabilities
        & \((0,0,1,0)\) \\
    \addlinespace[0.5ex]

    Critic goal-sampling probabilities
        & \raisebox{-0.45\height}{%
            \shortstack[l]{%
                \((0.2,0,0.5,0.3)\) (default)\\
                \((0,1,0,0)\) for TRL
            }} \\

    \bottomrule
    \end{tabular}
    \caption{Shared hyperparameters for the baseline algorithms.
    \(\dagger\) OTA uses a batch size of 1024 for AntMaze tasks and 256 for all other tasks.}
    \label{tab:appendix_baseline_shared}
\end{table}

\paragraph{OTA.}
Table~\ref{tab:appendix_ota_params} reports the environment-specific policy weights, temporal-abstraction settings, and discount factors used for OTA.

\begin{table}[h]
\centering
\small
\setlength{\tabcolsep}{5pt}
\renewcommand{\arraystretch}{1.08}
\begin{tabular}{@{}lccccc@{}}
\toprule
\textbf{Environment} & \(\beta_H\) & \(\beta_L\) & \(K\) & \(n\) & \(\gamma\) \\
\midrule
\texttt{amm}   & 3.0 & 3.0 & 25 & 5 & 0.99 \\
\texttt{aml}   & 3.0 & 3.0 & 25 & 5 & 0.99 \\
\texttt{cs}    & 1.0 & 3.0 & 20 & 4 & 0.99 \\
\texttt{cd}    & 0.5 & 3.0 & 25 & 4 & 0.99 \\
\texttt{ct}    & 3.0 & 3.0 & 20 & 4 & 0.99 \\
\texttt{p3}    & 0.5 & 3.0 & 25 & 4 & 0.99 \\
\texttt{p4}    & 1.0 & 3.0 & 20 & 4 & 0.99 \\
\texttt{scene} & 1.0 & 3.0 & 10 & 4 & 0.99 \\
\bottomrule
\end{tabular}
\caption{Environment-specific OTA hyperparameters.}
\label{tab:appendix_ota_params}
\end{table}

\paragraph{TRL.}
Table~\ref{tab:appendix_trl_params} reports the environment-specific weighting, regularization, and discount settings used for TRL.

\begin{table}[t]
\centering
\small
\setlength{\tabcolsep}{5pt}
\renewcommand{\arraystretch}{1.08}
\begin{tabular}{@{}lcccc@{}}
\toprule
\textbf{Environment} & \(\alpha\) & \(\lambda\) & \(\kappa\) & \(\gamma\) \\
\midrule
\texttt{amm}   & 0.7  & 0.0 & 0.7 & 0.99 \\
\texttt{aml}   & 0.7  & 0.0 & 0.7 & 0.99 \\
\texttt{cs}    & 1.0  & 0.7 & 0.7 & 0.99 \\
\texttt{cd}    & 10.0 & 1.0 & 0.7 & 0.99 \\
\texttt{ct}    & 1.0  & 1.0 & 0.7 & 0.99 \\
\texttt{p3}    & 2.0  & 0.5 & 0.7 & 0.99 \\
\texttt{p4}    & 2.0  & 2.0 & 0.7 & 0.99 \\
\texttt{scene} & 1.0  & 1.0 & 0.7 & 0.99 \\
\bottomrule
\end{tabular}
\caption{Environment-specific TRL hyperparameters.}
\label{tab:appendix_trl_params}
\end{table}

\paragraph{DQC.}
Across all environments, DQC uses 32 best-of-\(N\) samples, expectile
distillation, quantile implicit backups, and 10 flow steps.
Table~\ref{tab:appendix_dqc_params} reports the environment-specific execution horizon \(h_a\), critic horizon \(h\), and implicit-backup parameters.

\begin{table}[h!]
\centering
\small
\setlength{\tabcolsep}{5pt}
\renewcommand{\arraystretch}{1.08}
\begin{tabular}{@{}lccccc@{}}
\toprule
\textbf{Environment} & \(h_a\) & \(h\) & \(\kappa_b\) & \(\kappa_d\) & \(\gamma\) \\
\midrule
\texttt{amm}   & 1 & 25 & 0.5  & 0.8 & 0.999 \\
\texttt{aml}   & 1 & 25 & 0.5  & 0.8 & 0.999 \\
\texttt{cs}    & 1 & 5  & 0.99 & 0.8 & 0.999 \\
\texttt{cd}    & 1 & 5  & 0.99 & 0.8 & 0.999 \\
\texttt{ct}    & 5 & 25 & 0.7  & 0.8 & 0.999 \\
\texttt{p3}    & 5 & 10 & 0.7  & 0.8 & 0.999 \\
\texttt{p4}    & 1 & 5  & 0.7  & 0.5 & 0.999 \\
\texttt{scene} & 1 & 5  & 0.9  & 0.5 & 0.999 \\
\bottomrule
\end{tabular}
\caption{Environment-specific DQC hyperparameters.}
\label{tab:appendix_dqc_params}
\end{table}

\paragraph{Baseline hyperparameter tuning.}
For TRL, we tuned \(\alpha\in\{0.1,0.3,0.7\}\) on \texttt{amm} while
fixing \(\lambda=0\). On \texttt{ct}, we jointly tuned
\(\alpha\in\{1,10\}\) and \(\lambda\in\{0.7,1\}\). For OTA, we tuned
\(\beta_H\in\{0.5,1,3\}\) and \(K\in\{20,25\}\) on the Cube and Puzzle
tasks while fixing \(\beta_L=3\) and \(n=4\). On \texttt{scene}, we tuned
\(\beta_H\in\{0.5,1,3\}\) and \(K\in\{10,20\}\), again fixing
\(\beta_L=3\) and \(n=4\). The selected settings are reported in
Tables~\ref{tab:appendix_ota_params} and~\ref{tab:appendix_trl_params}.

\section{Ablation and Diagnostic Analyses}
\label{app:ablations}

The following analyses separate endpoint proposal, endpoint selection, and
bridge-based execution. Training ablations are retrained from scratch with the
main four-seed protocol. Inference-only ablations reuse the same trained
checkpoints and evaluation episode seeds; candidate samples are matched when
applicable to reduce variation unrelated to the analyzed component.

\subsection{Bridge-Removal Ablation}

HTVL retains PBG's transitive value learner, Gaussian endpoint proposer, and endpoint-selection procedure, but replaces bridge-based execution with HIQL's endpoint-conditioned low-level actor. HTVL improves the overall average from \(33.8\) for HIQL to \(46.0\), while PBG further reaches \(60.7\). Since HTVL and PBG share the high-level endpoint pipeline, the \(14.7\)-point gap supports the contribution of bridge construction and inverse-dynamics decoding.

\subsection{Endpoint Proposal and Selection}
\label{app:proposal_selection_ablations}

\paragraph{PBF selection-score decomposition.}
We evaluate the endpoint-selection rule on the six environments whose final
PBF configuration uses multiple endpoint candidates (\(N>1\)), for which
candidate ranking is nontrivial. We focus on PBF because several tasks benefit
from broader candidate sampling, whereas PBG generally favors deterministic
or low-breadth sampling and does not consistently improve as the candidate
count increases. We therefore omit the corresponding PBG analysis. Using the
main environment-specific \((N,T)\) settings and the same sampled candidate
sets across all selection rules, we compare the full transitive score
\(S_{\mathrm{tr}}(s,z,g)=V(s,z)V(z,g)\) against continuation-only ranking
with \(V(z,g)\) and uniform candidate selection.

\begin{table}[t]
\centering
\small
\setlength{\tabcolsep}{3.5pt}
\renewcommand{\arraystretch}{1.08}
\begin{tabular}{@{}lccc@{}}
\toprule
\textbf{Environment} & \(S_{\mathrm{tr}}(s,z,g)\) & \(V(z,g)\) & Uniform \\
\midrule
\texttt{amm}   & \(\mathbf{96.4 \pm 0.6}\) & \(95.8 \pm 0.5\) & \(95.1 \pm 1.4\) \\
\texttt{aml}   & \(\mathbf{84.6 \pm 6.4}\) & \(83.1 \pm 5.9\) & \(71.1 \pm 7.0\) \\
\texttt{cd}    & \(\mathbf{85.1 \pm 2.2}\) & \(83.7 \pm 2.4\) & \(82.4 \pm 2.1\) \\
\texttt{p3}    & \(\mathbf{69.3 \pm 3.3}\) & \(61.4 \pm 5.7\) & \(6.9 \pm 1.9\) \\
\texttt{p4}    & \(\mathbf{78.1 \pm 6.0}\) & \(65.1 \pm 9.0\) & \(4.4 \pm 0.5\) \\
\texttt{scene} & \(\mathbf{69.1 \pm 3.0}\) & \(68.6 \pm 3.8\) & \(56.0 \pm 4.0\) \\
\midrule
\textbf{Average} & \(\mathbf{80.4}\) & \(76.3\) & \(52.6\) \\
\bottomrule
\end{tabular}
\caption{Endpoint-selection-score ablation using the final trained proposer. Values are mean \(\pm\) std over four seeds.}
\label{tab:selection_score_ablation}
\end{table}

\subsection{Oracle-Waypoint Diagnostic on Puzzle}
\label{app:oracle_waypoints}

We compute a valid sequence of board configurations using the standard
linear-algebraic solution of Lights Out~\citep{anderson1998turning}. Each
waypoint is supplied in one of two roles. When used as the current task goal,
the long-horizon problem is decomposed into shorter goal-reaching problems
while learned subgoal proposal and selection remain active.
When supplied directly as the selected subgoal, learned proposal and selection
are bypassed, leaving only execution.

As reported in the main paper, performance improves markedly when the waypoint is supplied as the selected subgoal. In this condition, PBG executes the oracle sequence
with near-perfect success on both \texttt{p3} and \texttt{p4}. The gap between
the two waypoint roles therefore suggests that learned subgoal proposal and
selection remain a key bottleneck on Puzzle. Under the same selected subgoals,
PBG also substantially outperforms HIQL, supporting the contribution of
bridge-based execution over endpoint-conditioned low-level control.

\section{Hyperparameter Sensitivity}
\label{app:sensitivity}

We examine three PathBridger hyperparameters: the endpoint-sampling
configuration \((N,T)\), bridge horizon \(K\), and endpoint-weighting scale
\(c_{\rm sg}\). For each sweep, all remaining settings are fixed within each
environment and variant. The \((N,T)\) sweep changes only inference-time
sampling and reuses the trained checkpoints, whereas \(K\) and \(c_{\rm sg}\)
affect training and therefore use separately trained models. Bold indicates
the setting used in the main benchmark; because the final configurations are
selected jointly, the bold entry need not be the largest value in every
individual sweep.

\subsection{Endpoint-Sampling Sweep}
\label{app:sweep_nt}

We evaluate four paired \((N,T)\) configurations,
\[
(1,0),\quad (2,0.25),\quad (16,0.5),\quad (32,1),
\]
ordered by increasing inference-time sampling breadth. Here, \(N\) is the
number of endpoint candidates and \(T\) controls sampling dispersion. The
\((1,0)\) setting produces one deterministic endpoint: the Gaussian mean for
PBG and the zero-source flow endpoint for PBF. At the other extreme,
\((32,1)\) uses full-temperature sampling with 32 candidates. Because \(N\)
and \(T\) vary jointly, this sweep characterizes the overall breadth of the
candidate set rather than the isolated effect of either parameter.

\begin{table}[t]
\centering
\small
\setlength{\tabcolsep}{1.8pt}
\renewcommand{\arraystretch}{1.10}
\begin{tabular}{@{}lcccc@{}}
\toprule
\textbf{Environment} & \((1,0)\) & \((2,0.25)\) & \((16,0.5)\) & \((32,1)\) \\
\midrule
\texttt{amm}   & \(94.4 \pm 0.6\) & \(\mathbf{96.4 \pm 0.6}\) & \(93.5 \pm 0.2\) & \(88.9 \pm 2.0\) \\
\texttt{aml}   & \(79.7 \pm 4.5\) & \(80.9 \pm 3.3\) & \(\mathbf{84.6 \pm 6.4}\) & \(76.2 \pm 9.7\) \\
\texttt{cs}    & \(\mathbf{99.4 \pm 0.5}\) & \(96.9 \pm 0.4\) & \(92.3 \pm 1.0\) & \(63.7 \pm 4.1\) \\
\texttt{cd}    & \(86.8 \pm 7.6\) & \(\mathbf{85.1 \pm 2.2}\) & \(84.8 \pm 8.9\) & \(71.4 \pm 10.2\) \\
\texttt{ct}    & \(\mathbf{79.8 \pm 4.5}\) & \(78.2 \pm 1.6\) & \(76.2 \pm 1.3\) & \(58.1 \pm 0.8\) \\
\texttt{p3}    & \(17.0 \pm 3.3\) & \(21.1 \pm 3.7\) & \(54.1 \pm 3.7\) & \(\mathbf{69.3 \pm 3.3}\) \\
\texttt{p4}    & \(19.5 \pm 3.8\) & \(27.2 \pm 5.4\) & \(70.6 \pm 3.5\) & \(\mathbf{78.1 \pm 6.0}\) \\
\texttt{scene} & \(63.1 \pm 4.9\) & \(64.9 \pm 4.2\) & \(\mathbf{69.1 \pm 3.0}\) & \(57.7 \pm 6.4\) \\
\bottomrule
\end{tabular}
\caption{PBF success rates across the joint \((N,T)\) sweep. Values are mean \(\pm\) one standard deviation over four seeds. Bold indicates the setting used in the main benchmark.}
\label{tab:nt_sweep_min4_pbf}
\end{table}

\begin{table}[t]
\centering
\small
\setlength{\tabcolsep}{1.8pt}
\renewcommand{\arraystretch}{1.10}
\begin{tabular}{@{}lcccc@{}}
\toprule
\textbf{Environment} & \((1,0)\) & \((2,0.25)\) & \((16,0.5)\) & \((32,1)\) \\
\midrule
\texttt{amm}   & \(\mathbf{95.3 \pm 0.6}\) & \(93.3 \pm 2.1\) & \(91.0 \pm 0.7\) & \(79.5 \pm 2.3\) \\
\texttt{aml}   & \(\mathbf{85.3 \pm 3.0}\) & \(81.8 \pm 4.0\) & \(75.1 \pm 4.0\) & \(60.5 \pm 3.1\) \\
\texttt{cs}    & \(\mathbf{93.0 \pm 1.7}\) & \(79.1 \pm 4.1\) & \(67.1 \pm 4.0\) & \(28.7 \pm 3.7\) \\
\texttt{cd}    & \(\mathbf{75.7 \pm 1.9}\) & \(58.9 \pm 4.2\) & \(43.6 \pm 4.9\) & \(16.6 \pm 4.3\) \\
\texttt{ct}    & \(\mathbf{45.1 \pm 5.8}\) & \(29.7 \pm 1.4\) & \(22.1 \pm 1.6\) & \(11.0 \pm 0.8\) \\
\texttt{p3}    & \(25.8 \pm 5.7\) & \(\mathbf{28.9 \pm 3.6}\) & \(26.3 \pm 3.7\) & \(12.4 \pm 0.6\) \\
\texttt{p4}    & \(1.5 \pm 1.1\) & \(1.7 \pm 1.0\) & \(\mathbf{2.7 \pm 1.6}\) & \(0.6 \pm 0.2\) \\
\texttt{scene} & \(39.9 \pm 12.4\) & \(45.5 \pm 5.8\) & \(\mathbf{59.5 \pm 3.8}\) & \(36.9 \pm 4.3\) \\
\bottomrule
\end{tabular}
\caption{PBG success rates across the joint \((N,T)\) sweep. Values are mean \(\pm\) one standard deviation over four seeds. Bold indicates the setting used in the main benchmark.}
\label{tab:nt_sweep_min4_pbg}
\end{table}

The preferred sampling breadth is task-dependent. PBG generally favors narrow
sampling, with \((1,0)\) performing best on five of the eight environments.
PBF also favors concentrated sampling on \texttt{amm} and the Cube tasks,
where PBG is relatively competitive. In contrast, broader PBF sampling is
important on Puzzle: performance rises from \(17.0/19.5\) under \((1,0)\) to
\(69.3/78.1\) under \((32,1)\) on \texttt{p3}/\texttt{p4}. The
\texttt{aml} and \texttt{scene} tasks peak at the intermediate
\((16,0.5)\) setting. This pattern is consistent with concentrated sampling being sufficient on several tasks, whereas Puzzle benefits strongly from broader candidate coverage.

\subsection{Bridge-Horizon Sweep}
\label{app:sweep_k}

The bridge horizon \(K\) determines both the endpoint-prediction horizon and
the length of the generated state-space bridge. We compare
\(K\in\{25,40\}\) while fixing all non-\(K\) settings within each environment
and variant on a single seed.

\begin{table}[t]
\centering
\small
\setlength{\tabcolsep}{5.0pt}
\renewcommand{\arraystretch}{1.10}
\begin{tabular}{@{}lcccc@{}}
\toprule
& \multicolumn{2}{c}{\textbf{PBF}} & \multicolumn{2}{c}{\textbf{PBG}} \\
\cmidrule(lr){2-3}\cmidrule(lr){4-5}
\textbf{Environment} & \(K{=}25\) & \(K{=}40\) & \(K{=}25\) & \(K{=}40\) \\
\midrule
\texttt{amm}   & \textbf{97.3} & 91.5 & \textbf{96.0} & 87.7 \\
\texttt{aml}   & \textbf{93.3} & 83.5 & \textbf{89.1} & 82.1 \\
\texttt{cs}    & 95.7 & \textbf{97.1} & \textbf{94.9} & 87.2 \\
\texttt{cd}    & 80.4 & \textbf{82.4} & \textbf{77.3} & 64.3 \\
\texttt{ct}    & 55.5 & \textbf{77.3} & \textbf{42.4} & 6.9 \\
\texttt{p3}    & \textbf{68.5} & 55.7 & 15.5 & \textbf{33.9} \\
\texttt{p4}    & \textbf{82.1} & 56.5 & 0.0 & \textbf{4.3} \\
\texttt{scene} & \textbf{53.6} & 64.3 & 31.7 & \textbf{65.2} \\
\midrule
\textbf{Average} & 78.3 & 76.0 & 55.9 & 54.0 \\
\bottomrule
\end{tabular}
\caption{Success rates across the bridge-horizon sweep. All non-\(K\) settings are fixed within each environment and variant. Bold indicates the setting used in the main benchmark.}
\label{tab:controlled_k}
\end{table}

The preferred horizon is also task-dependent. PBF benefits from \(K=40\) on
all Cube tasks, with the largest increase on \texttt{ct}
(\(55.5\rightarrow77.3\)), whereas \(K=25\) performs better on AntMaze and
Puzzle. PBG generally favors \(K=25\) on AntMaze and Cube, while \(K=40\)
performs better on Puzzle and \texttt{scene}. Averaged across tasks,
\(K=25\) is slightly better for both PBF (\(78.3\) vs.\ \(76.0\)) and PBG
(\(55.9\) vs.\ \(54.0\)). The contrasting Cube trend is consistent with the
flow proposer handling the endpoint distributions induced by longer horizons
more effectively than the Gaussian proposer.

\subsection{Endpoint-Weighting Sweep}
\label{app:sweep_csg}

The endpoint-weighting scale \(c_{\rm sg}\) controls how strongly the
continuation-value improvement tilts the empirical endpoint distribution. We
compare \(c_{\rm sg}\in\{5,10\}\) while fixing all remaining settings within
each environment and variant on a single seed.

\begin{table}[t]
\centering
\small
\setlength{\tabcolsep}{4.2pt}
\renewcommand{\arraystretch}{1.10}
\begin{tabular}{@{}lcccc@{}}
\toprule
& \multicolumn{2}{c}{\textbf{PBF}} & \multicolumn{2}{c}{\textbf{PBG}} \\
\cmidrule(lr){2-3}\cmidrule(lr){4-5}
\textbf{Environment} & \(c_{\rm sg}=5\) & \(c_{\rm sg}=10\) & \(c_{\rm sg}=5\) & \(c_{\rm sg}=10\) \\
\midrule
\texttt{amm}   & 94.9 & \textbf{97.3} & 91.2 & \textbf{95.7} \\
\texttt{aml}   & 78.1 & \textbf{78.4} & 70.1 & \textbf{82.1} \\
\texttt{cs}    & \textbf{99.7} & 98.9 & 83.2 & \textbf{92.5} \\
\texttt{cd}    & 82.9 & \textbf{87.9} & 55.7 & \textbf{77.1} \\
\texttt{ct}    & 74.1 & \textbf{77.3} & 33.1 & \textbf{42.4} \\
\texttt{p3}    & 64.3 & \textbf{68.5} & 21.3 & \textbf{33.9} \\
\texttt{p4}    & 76.3 & \textbf{82.1} & 0.3 & \textbf{4.3} \\
\texttt{scene} & \textbf{70.9} & 48.5 & \textbf{57.6} & 54.8 \\
\midrule
\textbf{Average} & 80.2 & 79.9 & 51.6 & 60.4 \\
\bottomrule
\end{tabular}
\caption{Success rates across the endpoint-weighting sweep. All non-\(c_{\rm sg}\) settings are fixed within each environment and variant. Bold indicates the setting used in the main benchmark.}
\label{tab:controlled_csg}
\end{table}

A stronger value tilt generally benefits PBG: \(c_{\rm sg}=10\) improves
seven of the eight environments and raises the average from \(51.6\) to
\(60.4\). PBF is less sensitive in aggregate (\(80.2\) vs.\ \(79.9\)), but
its preferred scale remains task-dependent. The stronger tilt improves most
AntMaze, Cube, and Puzzle settings, whereas \(c_{\rm sg}=5\) performs better
on \texttt{cs} and especially \texttt{scene}. Thus, stronger value weighting is generally useful for the Gaussian proposer.
PBF is less sensitive in aggregate, although individual tasks, most notably
\texttt{scene}, remain sensitive to the endpoint-weighting scale.

    
\section{Additional Results and Diagnostics}
\label{app:additional_diagnostics}

\subsection{Full Data-Scaling Results}
\label{app:data_scaling_full}

The main paper reports the aggregate trend across dataset sizes.
Table~\ref{tab:data_scaling_full} provides the available per-environment
results at 100k, 300k, 500k, and 1M transitions. All methods are trained for 1M
gradient steps at each data budget. Values are mean \(\pm\) one standard
deviation over two seeds.

\begin{table*}[b!]
    \centering
    \small
    \setlength{\tabcolsep}{1.7pt}
    \renewcommand{\arraystretch}{1.08}
    \begin{tabular}{@{}clccccccccc@{}}
    \toprule
    \textbf{Transitions} & \textbf{Method} & \texttt{amm} & \texttt{aml} & \texttt{cs} & \texttt{cd} & \texttt{ct} & \texttt{p3} & \texttt{p4} & \texttt{scene} & \textbf{Average} \\
    \midrule
    \multirow{5}{*}{100k}
    & HIQL & $24.6\pm0.1$ & $12.3\pm1.4$ & $4.9\pm1.7$ & $0.2\pm0.2$ & $0.1\pm0.1$ & $1.7\pm0.3$ & $0.0\pm0.0$ & $4.5\pm1.4$ & $6.0$ \\
    & TRL & $9.2\pm2.8$ & $0.0\pm0.0$ & $\mathbf{21.4\pm0.1}$ & $0.1\pm0.2$ & $0.1\pm0.2$ & $1.5\pm0.5$ & $0.1\pm0.1$ & $2.4\pm3.3$ & $4.3$ \\
    & DQC & $18.4\pm0.0$ & $5.7\pm0.0$ & $13.9\pm0.0$ & $0.5\pm0.0$ & $0.0\pm0.0$ & $\mathbf{56.4\pm0.0}$ & $0.5\pm0.0$ & $10.8\pm0.0$ & $13.3$ \\
    & PBG & $81.2\pm12.0$ & $28.6\pm6.6$ & $10.2\pm2.1$ & $0.7\pm0.4$ & $0.0\pm0.0$ & $5.3\pm4.7$ & $0.0\pm0.0$ & $4.1\pm0.5$ & $16.3$ \\
    & PBF & $\mathbf{87.4\pm2.6}$ & $\mathbf{37.5\pm4.2}$ & $\mathbf{22.4\pm0.4}$ & $\mathbf{5.4\pm0.3}$ & $\mathbf{0.3\pm0.3}$ & $38.0\pm4.0$ & $\mathbf{34.8\pm0.1}$ & $\mathbf{14.3\pm4.5}$ & $\mathbf{30.0}$ \\
    \midrule
    \multirow{5}{*}{300k}
    & HIQL & $90.4\pm1.2$ & $\mathbf{80.5\pm2.1}$ & $4.6\pm0.2$ & $0.6\pm0.4$ & $0.0\pm0.0$ & $2.5\pm0.6$ & $0.1\pm0.2$ & $11.6\pm2.7$ & $23.8$ \\
    & TRL & $55.5\pm1.4$ & $11.6\pm0.1$ & $46.3\pm1.4$ & $0.7\pm0.3$ & $0.1\pm0.1$ & $17.4\pm0.4$ & $3.1\pm0.8$ & $11.9\pm3.4$ & $18.3$ \\
    & DQC & $88.0\pm1.6$ & $68.5\pm7.9$ & $27.8\pm1.4$ & $4.0\pm2.8$ & $0.0\pm0.0$ & $\mathbf{87.1\pm1.1}$ & $\mathbf{98.5\pm0.6}$ & $55.4\pm4.7$ & $53.7$ \\
    & PBG & $\mathbf{91.3\pm0.6}$ & $61.2\pm2.1$ & $\mathbf{69.8\pm6.6}$ & $5.1\pm0.9$ & $0.1\pm0.1$ & $14.0\pm1.7$ & $0.1\pm0.1$ & $15.8\pm2.0$ & $32.2$ \\
    & PBF & $\mathbf{93.8\pm0.7}$ & $72.0\pm0.6$ & $\mathbf{73.3\pm2.0}$ & $\mathbf{37.0\pm0.8}$ & $\mathbf{12.6\pm1.6}$ & $53.3\pm0.6$ & $48.0\pm5.2$ & $\mathbf{64.1\pm1.7}$ & $\mathbf{56.8}$ \\
    \midrule
    \multirow{5}{*}{500k}
    & HIQL & $\mathbf{93.5\pm1.4}$ & $\mathbf{87.2\pm0.7}$ & $9.4\pm2.1$ & $1.2\pm0.1$ & $0.1\pm0.2$ & $2.1\pm0.3$ & $0.2\pm0.1$ & $25.0\pm1.0$ & $27.3$ \\
    & TRL & $81.9\pm5.7$ & $24.1\pm0.0$ & $62.4\pm6.4$ & $0.9\pm0.1$ & $0.2\pm0.4$ & $79.1\pm1.1$ & $7.2\pm10.2$ & $15.5\pm20.7$ & $33.9$ \\
    & DQC & $82.6\pm3.1$ & $68.0\pm6.6$ & $35.1\pm0.3$ & $5.0\pm0.3$ & $0.3\pm0.1$ & $\mathbf{90.4\pm13.4}$ & $\mathbf{99.9\pm0.0}$ & $\mathbf{68.8\pm8.5}$ & $56.3$ \\
    & PBG & $\mathbf{89.4\pm0.1}$ & $80.8\pm6.7$ & $85.9\pm0.0$ & $35.0\pm9.8$ & $0.1\pm0.1$ & $16.1\pm1.3$ & $0.1\pm0.0$ & $23.4\pm11.7$ & $41.3$ \\
    & PBF & $\mathbf{89.2\pm6.2}$ & $\mathbf{83.0\pm3.0}$ & $\mathbf{93.2\pm6.9}$ & $\mathbf{73.9\pm9.7}$ & $\mathbf{35.2\pm2.3}$ & $56.2\pm6.9$ & $56.2\pm4.9$ & $38.6\pm29.8$ & $\mathbf{65.7}$ \\
    \midrule
    \multirow{5}{*}{1M}
    & HIQL & $\mathbf{95.8\pm0.3}$ & $\mathbf{88.1\pm5.2}$ & $15.4\pm1.6$ & $6.8\pm0.1$ & $2.3\pm1.4$ & $9.4\pm0.8$ & $8.6\pm3.8$ & $44.6\pm2.0$ & $33.9$ \\
    & TRL & $87.2\pm1.2$ & $48.5\pm10.3$ & $92.8\pm3.8$ & $22.5\pm1.7$ & $14.9\pm0.6$ & $\mathbf{99.8\pm0.1}$ & $33.8\pm4.2$ & $70.3\pm0.1$ & $58.7$ \\
    & DQC & $87.6\pm1.3$ & $70.9\pm8.5$ & $75.2\pm3.4$ & $16.9\pm1.8$ & $32.3\pm1.4$ & $\mathbf{100.0\pm0.0}$ & $\mathbf{100.0\pm0.0}$ & $\mathbf{93.1\pm3.0}$ & $72.0$ \\
    & PBG & $\mathbf{95.2\pm0.4}$ & $83.2\pm1.6$ & $93.7\pm1.7$ & $75.2\pm2.7$ & $44.4\pm2.8$ & $29.6\pm6.1$ & $2.4\pm2.7$ & $58.0\pm0.6$ & $60.2$ \\
    & PBF & $\mathbf{96.4\pm1.0}$ & $81.6\pm4.5$ & $\mathbf{99.1\pm0.6}$ & $\mathbf{85.2\pm3.9}$ & $\mathbf{81.8\pm6.4}$ & $67.7\pm1.1$ & $82.1\pm0.0$ & $68.9\pm2.8$ & $\mathbf{82.8}$ \\
    \bottomrule
    \end{tabular}
    \caption{Per-environment dataset-size results. Values are mean \(\pm\) one standard deviation over two seeds. All methods are trained for 1M gradient steps at each data budget. Bold indicates mean performance at or above 95\% of the best mean in each setting.}
    \label{tab:data_scaling_full}
\end{table*}

\end{document}